\documentclass[lettersize,journal]{IEEEtran}
\usepackage{amsmath,amsfonts}
\usepackage{algorithmic}
\usepackage{algorithm}
\usepackage{array}
\usepackage[caption=false,font=normalsize,labelfont=sf,textfont=sf]{subfig}
\usepackage{textcomp}
\usepackage{stfloats}
\usepackage{url}
\usepackage{verbatim}
\usepackage{graphicx}
\usepackage{cite}
\usepackage{xspace}
\usepackage{multirow}
\usepackage{nicefrac}       
\usepackage{microtype}      
\usepackage{xcolor}         
\usepackage{amssymb}
\usepackage{booktabs}
\usepackage{wrapfig}
\usepackage{capt-of}

\def\eg{\emph{e.g.}} 
\def\ie{\emph{i.e.}}

\newcommand{\PreserveBackslash}[1]{\let\temp=\\#1\let\\=\temp}
\newcolumntype{C}[1]{>{\PreserveBackslash\centering}p{#1}}
\newcolumntype{R}[1]{>{\PreserveBackslash\raggedleft}p{#1}}
\newcolumntype{L}[1]{>{\PreserveBackslash\raggedright}p{#1}}

\begin{document}

\title{In Defense of Clip-based Video Relation Detection}

\author{Meng Wei, Long Chen, Wei Ji, Xiaoyu Yue, Roger Zimmermann
        \thanks{Meng Wei is with the Department of Electrical and Electronic Engineering, The University of Hong Kong, Pokfulam, Hong Kong SAR, 999077. 
        Email: mengwei.kelly@connect.hku.hk}
        \thanks{Long Chen is with the Department of Computer Science and Engineering, The Hong Kong University of Science and Technology, Kowloon, Hong Kong SAR, 999077. Email: longchen@ust.hk.}
        \thanks{Wei Ji and Roger Zimmermann are with the School of Computing, National University of Singapore, Singapore, 117417. Email: jiwei@nus.edu.sg; rogerz@comp.nus.edu.sg.}
        \thanks{Xiaoyu Yue is with the School of Electrical and Information Engineering, The University of Sydney, Camperdown NSW, Australia, 2006. Email: yuexiaoyu002@gmail.com}
        
}

\markboth{IEEE Transactions on Image Processing}{}


\maketitle


\begin{abstract}
Video Visual Relation Detection (VidVRD) aims to detect visual relationship triplets in videos using spatial bounding boxes and temporal boundaries.
Existing VidVRD methods can be broadly categorized into bottom-up and top-down paradigms, depending on their approach to classifying relations.
Bottom-up methods follow a clip-based approach where they classify relations of short clip tubelet pairs\footnote{We use ``clip tubelets'' and ``video tubelets'' to denote the tubelets generated from clips or video, respectively. \label{tube}} and then merge them into long video relations. On the other hand, top-down methods directly classify long video tubelet pairs. 
While recent video-based methods utilizing video tubelets have shown promising results, we argue that the effective modeling of spatial and temporal context plays a more significant role than the choice between clip tubelets and video tubelets.
This motivates us to revisit the clip-based paradigm and explore the key success factors in VidVRD.
In this paper, we propose a Hierarchical Context Model (HCM) that enriches the \textbf{object-based spatial context} and \textbf{relation-based temporal context} based on clips. We demonstrate that using clip tubelets can achieve superior performance compared to most video-based methods. Additionally, using clip tubelets offers more flexibility in model designs and helps alleviate the limitations associated with video tubelets, such as the challenging long-term object tracking problem and the loss of temporal information in long-term tubelet feature compression.
Extensive experiments conducted on two challenging VidVRD benchmarks validate that our HCM achieves a new state-of-the-art performance, highlighting the effectiveness of incorporating advanced spatial and temporal context modeling within the clip-based paradigm.
\end{abstract}


\section{Introduction}
\label{sec:intro}
\begin{figure*}
    \centering
    \begin{minipage}{1.0\textwidth}
        \centering
        \includegraphics[width=1.0\textwidth]{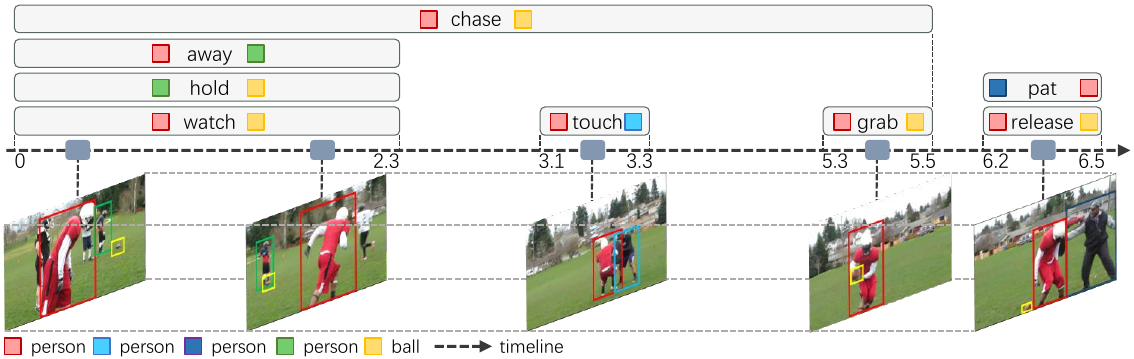}
        (a)
    \end{minipage}
    \begin{minipage}{1.0\textwidth}
        \centering
        \vspace{0.01\linewidth}
        \includegraphics[width=1.0\textwidth]{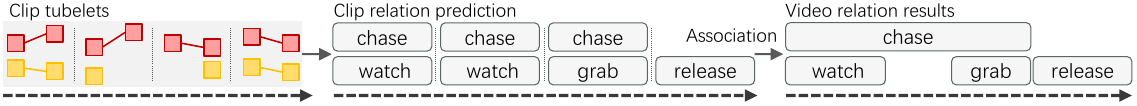}
        (b)
    \end{minipage}
    \begin{minipage}{1.0\textwidth}
        \centering
        \includegraphics[width=1.0\textwidth]{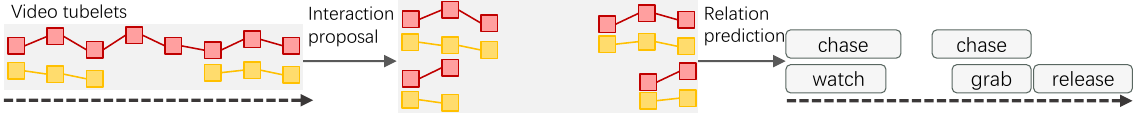}
        (c)
    \end{minipage}
    \caption{(a) An illustration of video relations in the event of a rugby match. Take the video relations between \textcolor{red}{person} and \textcolor{yellow}{ball} for example, (b) is the pipeline of the bottom-up clip-based approaches, (c) is the pipeline of the top-down video-based approaches.}
    \label{fig:intro}
\end{figure*}

\IEEEPARstart{W}{ith} the advancement of object tracking~\cite{xu2019learning,danelljan2015convolutional,zhu2018online,wojke2017simple,chen2018real} and action recognition~\cite{wang2016temporal,peng2016bag,wang2018non,yan2018spatial,tran2018closer}, dynamic scene understanding should go deeper into recognizing the object interactions (or relations) throughout time.
In order to achieve this, Video Visual Relation Detection (VidVRD) has been proposed, which aims to detect visual relationship triplets ``\texttt{subject-predicate-object}'' in videos (\eg, \texttt{person-chase-ball}) and localize the corresponding spatial and temporal boundaries for the involved entities (\ie, object tubelets).
Modeling the complex and time-varying interactions among object tubelets is challenging.
This is because short-term and long-term visual relations have different emphases on the local and global spatial-temporal clues.
Figure~\ref{fig:intro} (a) provides an illustration of this distinction. 
The short-term predicates, such as "grab" and "pat," primarily rely on fine-grained spatial clues. 
Conversely, long-term predicates like "away" and "chase" depend on global temporal clues.

Most existing VidVRD methods employ object tubelets generated by off-the-shelf tracking algorithms to model pairwise relation features.
The early proposed methods~\cite{shang2017video,tsai2019video,qian2019video,ji2021vidvrd} adopt a bottom-up clip-based approach.
As shown in Figure~\ref{fig:intro} (b), they divide the videos into clips and generate short-term tubelets to predict initial relations. 
These clip relations are then merged into video relations using a relation association algorithm.
In contrast, recent video-based methods~\cite{liu2020beyond,chen2021social,gao2021classification} have deviated from the use of short-term clip tubelets.
They instead focused on enhancing long-term temporal dependencies by generating complete video object trajectories. 
Following a top-down approach, as shown in Figure~\ref{fig:intro} (c), these methods~\cite{liu2020beyond,chen2021social} begin by designing a relation proposal stage. In this stage, they sample both short-term and long-term video tubelet pairs from the entire object trajectories for a comprehensive relation candidate set.
Subsequently, a prediction stage is employed to classify the relations of the sampled tubelet pairs.
An alternative method~\cite{gao2021classification} firstly classifies all the short-term and long-term relations based on the entire video tubelets, without explicitly proposing relation candidates. 
Then the the temporal boundaries of the classified relations are grounded in the second stage.

Overall, the state-of-the-art video-based methods mainly argue that using clip tubelets are insufficient in capturing long-term spatial-temporal context. 
And using long-term video tubelets leads to improved performance particularly in handling long-term relationships. 
Hereby, the video-based paradigm which relies on video tubelets has emerged as the de-facto standard in VidVRD recently.
However, we note that using video tubelets for VidVRD also presents its own challenge.
The generation of a feasible set of video tubelets often relies on learning temporal priors specific to certain relationships. 
For example, BeyondShort~\cite{liu2020beyond} learns an individual relationship pair proposal network to filter out noisy pairs generated by the multi-scale sliding window scheme. 
Similarly, SocialFab~\cite{chen2021social} trains a binary classifier to learn per-frame interactiveness scores of each object pair. Then they employ the watershed algorithm to generate relation pair proposals based on the interactiveness scores. 
However, explicitly learning to propose temporal tubelets introduces three notable limitations:
a) The generalization of video tubelet proposals tends to be weak across different datasets.
b) The inclusion of multiple stages in the pipeline increases complexity.
c) The computation cost is comparable to or even higher than clip-based methods.

In this paper, we contend that the performance disparity between clip-based and video-based methods is primarily a result of the specific techniques employed for temporal modeling, rather than the inferiority of clip tubelets. 
More importantly, we find that by incorporating advanced temporal modeling into clip-based methods, it is possible to narrow the performance gap and even surpass the state-of-the-art video-based methods.
Accordingly, we enhance the original clip-based paradigm in two aspects: 
(1) \textit{Spatial Enhancement}:
As illustrated in Figure~\ref{fig:intro}, the relations between \texttt{persons} and \texttt{ball} are highly predictive with their co-occurrence and highly relevant to their relative positions. 
Hence, instead of considering individual objects, we incorporate the group-level cues (i.e., the relative spatial arrangement and object co-occurrence) that have been proven effective in various computer vision tasks, including object recognition~\cite{galleguillos2008object}, object detection~\cite{chen2017spatial,hu2018relation} and scene graph generation~\cite{woo2018linknet, chen2019knowledge}. 
(2) \textit{Temporal Enhancement}: Recognizing the temporal order is crucial in activity or event extraction~\cite{zhou2018temporal}. For example, ``\texttt{person-chase-ball}'', ``\texttt{person-grab-ball}'', and ``\texttt{person-release-ball}'' always sequentially describe the main steps of a rugby match event.
Thereby, we further establish the correspondences between video relations along the ``\textit{arrow of the time}''. 

To achieve these goals, we propose a novel Hierarchical Context Model (HCM) based on clip tubelets to learn multi-level contexts, namely from spatial dimension to temporal dimension, short-term to long-term, and low-order to high-order. 
\textbf{The first level is the object-based spatial context}. We construct a spatial affinity graph that consists of a \emph{positional} subgraph and a \emph{semantic} subgraph to capture the heterogeneity of relative spatial arrangement and object co-occurrence.
\textbf{The second level is the relation-based temporal context}. We leverage the clip index as the ``\textit{arrow of the time}'' to explicitly model the temporal correspondence of the continuous relations.
Due to the one-way direction nature of time, we design a unidirectional temporal affinity graph in which each vertex represents a clip tubelet pair and each edge describes the  correlations of adjacent relations in terms of a mixture of appearance affinity and location affinity.
As it is computationally expensive to learn the temporal connections of a quadratic complexity, we only keep the edges with high appearance affinity and location affinity.

Empowered by the hierarchical spatial-temporal context, HCM achieves state-of-the-art performance in VidVRD.
Additionally, our research verify that clip tubelets should not be a secondary choice in VidVRD, as they offer unique advantages over video tubelets.
One key advantage of using clip tubelets is their ability to mitigate the challenging long-term object tracking problem. The video-based approaches use an off-the-shelf tracker (e.g., DeepSORT) to generate video tubelets which compare objects of different frames and find their correspondence with clues such as appearance and motion.
The generated video tracklets often suffer from fragmentation due to occlusion, variations in quality etc.
However, the clip-based VidVRD models can provide additional clip-level clues in the form of ``\texttt{subject-predicate-object}'' triplets. 
This enables the recovery of complete tracklets when they are partially occluded in certain clips. For instance, if the subject ``\texttt{person}'' is occluded in some frames but the object ``\texttt{ball}'' remains visible, the subject's trajectory can be recovered in the relation association of the predicted long-term relation ``chase''.
On the other hand, using clip tubelets is more flexible in terms of the model design and training process. The full-length video tubelet feature is infeasible for training. Hence the video-based methods have to discard more temporal information (by pooling) in relation prediction. In order to determine the temporal boundary, they have to design multiple
training stages. Also, the spatial-temporal structure is hard to model in the video-based framework because the video tubelet pairs have different temporal lengths.  

\section{Related Work}
\begin{figure*}
    \centering
    \includegraphics[width=1.0\linewidth]{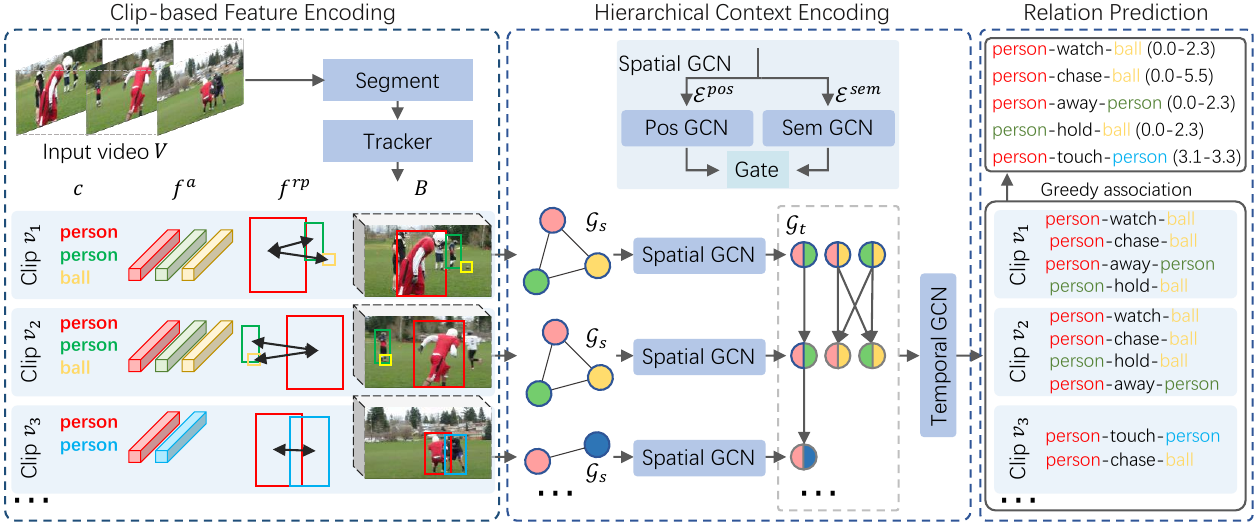}
    \caption{
    The Hierarchical Context Model (HCM) comprises three main modules: a clip-based feature encoding module, a hierarchical context encoding module, and a relation prediction module. To provide a clearer illustration, we specifically highlight the positive relation pairs in the temporal graph. \textbf{Clip-Based Feature Encoding Module}: extract visual representations from each individual object within the clip, including the visual appearances of objects and spatial relationship between two objects within the same clip.
    \textbf{Hierarchical Context Encoding Module}: incorporate spatial cues including relative spatial arrangement and object co-occurrence through two spatial GCNs at the object level and then establish the temporal correspondences of clip relations along the "arrow of time" through a temporal GCN at the relation level.
    \textbf{Relation Prediction Module}: classify both the short-term and long-term relations that each clip tubelet is involved in and link the clip-level relations to videl-level relations.
    }
    \label{fig:arch}
\end{figure*}

\noindent\textbf{Video Relation Detection (VidVRD).} 
VidVRD explores the temporal dynamics of video relations, aiming to capture the complexities of real-world activities.
Different from the image visual relation detection methods~\cite{xu2017scenegraph, Chen_2019_ICCV, li2017scene, zhang2022fine, wei2020hose, wei2022rethinking}, most VidVRD methods separately predict the relations between clip/video tubelet pairs because of the more complex spatial-temporal context structure. 
Mainstream methods in VidVRD can be categorized into clip-based and video-based approaches.
Most clip-based methods~\cite{shang2017video,tsai2019video, qian2019video,su2020video,li2021interventional,shang2021video} focus on designing multi-modality local features for clip tubelets to make short-term predictions. These methods excel at capturing spatial relationships within individual clips. However, they may struggle to model long-term dependencies and contextual information that span multiple clips.
Video-based methods~\cite{woo2021and,liu2020beyond, chen2021social,gao2021classification} 
address the limitations of clip-based methods by utilizing video tubelet features. These methods can predict long-term relations by considering the temporal coherence across multiple clips. However, they often face challenges in explicitly establishing the inter-dependencies between objects, short-term and long-term relations. Moreover, their performance heavily relies on accurate video tracking, which can be computationally expensive and prone to errors.

To overcome the limitations of the existing clip-based and video-based methods,
our proposed HCM focus on the following extensions:
(1) Integrating spatial and temporal context. We make efforts to incorporate both spatial and temporal context into the modeling process.
(2) Exploring long-term dependencies. We explicitly capture and model long-term dependencies between objects and their relations across multiple clips or frames. 

\noindent\textbf{GCN based VidVRD.}
Several clip-based methods~\cite{tsai2019video, qian2019video} have employed techniques such as Conditional Random Field~\cite{10.5555/645530.655813} or Graph Convolutional Networks (GCNs)~\cite{kipf2016semi,chiang2019cluster,chen2017stochastic,8809901,li2019deepgcns,hamilton2017inductive} to model spatial-temporal structures. 
For example, GSTEG~\cite{tsai2019video} fully connects objects and predicates across all time periods, but this approach becomes computationally expensive for long videos.
Similarly, VRD-GCN~\cite{qian2019video} connects objects in three consecutive clips (previous, current, and next) to avoid over-smoothing issues caused by dense long-term connections.
However, both methods utilize an integral spatial-temporal graph, which suffers from scalability issues.

The video-based method BeyondShort~\cite{liu2020beyond} also incorporates spatial and temporal graphs in its approach, but our HCM  differs significantly from BeyondShort in its design and utilization of these graphs.
In BeyondShort, the spatial graph and temporal graph are used only in the proposal stage, as their experiments demonstrate that the GCN feature does not provide benefits in the relation classification stage. This outcome is not surprising since BeyondShort's GCN model learns context through message passing among arbitrary video tubelets, which can lead to chaotic spatial and temporal context representation.
On the other hand, HCM takes a different approach by leveraging well-defined spatial-temporal structures with clip tubelets. Unlike BeyondShort, HCM focuses on learning temporal context at the relation-level rather than the object-level. This unique design allows HCM to capture meaningful temporal dependencies specific to the relationships being detected.
Furthermore, HCM incorporates graph and node affinity designs that are distinct from BeyondShort. These designs are tailored to the spatial and temporal context modeling within the clip-based paradigm, offering a more structured and coherent representation of spatial and temporal relationships.

\section{Method}
\subsection{Overview}
In this section, we give the overview of the proposed clip-based hierarchical framework HCM and summarize the notations used in following sections.
Given a video $V$, it is evenly segmented into $N$ overlapping clips with clip length $\delta$ (\ie, frame number). 
Each clip $v$ has $M$ clip tubelets.
The proposed HCM consists of three steps:
 (1)  We encode the clip-based features as introduced in Sec.~\ref{sec:clip_feat_enc}.
 (2) We first construct an object-based spatial graph $\mathcal{G}_s = \left\{\mathcal{V}_s, \mathcal{E}_s \right\}$ for each $v$, where $\mathcal{V}_s$ is the tubelet vertices and $\mathcal{E}_s$ is the spatial affinity matrix. 
Then a relation-based temporal graph $\mathcal{G}_t = \left\{\mathcal{V}_t, \mathcal{E}_t \right\}$ is constructed among all clips $V$, where $\mathcal{V}_t$ is the tubelet pair vertices and $\mathcal{E}_t$ is the temporal affinity matrix.
The encoding process is detailed in Sec.~\ref{sec:multi_level_context_enc}.
 (3) We first predict the clip-level relations and then use a relaxed greedy relation association algorithm to obtain the video-level relations with rough temporal boundaries as elaborated in Sec.~\ref{sec:relation_prediction}. 
The overall architecture is shown in Figure~\ref{fig:arch}.

\subsection{Clip-based Feature Encoding}
\label{sec:clip_feat_enc}
To represent complex relations between pairwise objects, we encode both the single object appearance feature and pairwise relative positional feature. \\
\textbf{Appearance Feature.} 
Following previous works~\cite{liu2020beyond, gao2021classification}, we adopt the CNN backbone of the pre-trained detector to extract the ROI pooling features for boxes in each tubelet. 
As it requires much memory to directly process all frames, we assume that the object appearance little changes in each clip's duration and hence we average all the box features of each tubelet as its appearance feature $f^a$.
\\
\textbf{Relative Spatial Feature.}
To complement the visual information, the relative position of the subject tubelet and object tubelet is regarded to be indispensable. 
We first compute the relative positional feature $f^{r}$ of two boxes in each frame:
\begin{equation}
\small
    f^{r} = \left[\frac{x_i-x_j}{wj}, \frac{y_i-y_j}{hj}, \log\frac{w_i}{w_j}, \log\frac{h_i}{h_j}, \log\frac{wi*hi}{wj*hj}\right]
\end{equation}
Then for computation efficiency, we concatenate $f^{r}$ of the start frame and $f^{r}$ of the end frame of each tubelet pair to get the relative spatial feature $f^{rp}$. 

\subsection{Hierarchical Spatial-Temporal Context}
\label{sec:multi_level_context_enc}
To model the multi-level spatial-temporal context, instead of using an integral spatial-temporal graph on objects, we split the video graph structure into an object-based spatial structure and a relation-based temporal structure. The reasons for this decoupling design are two-fold: 
(1) Coupling spatial-temporal message passing makes it hard to explicitly use different prior knowledge in two dimensions.
(2) Dense spatial-temporal connections easily cause the over-smoothing issue due to feature redundancy.
Hence, we incorporate the object-level knowledge into the spatial graph and the relation-level knowledge into the temporal graph. All details are presented in Sec.~\ref{sec:object_spatial_graph} and Sec.~\ref{sec:relation_temporal_graph}.  

\subsubsection{Object-based Spatial Graph Construction}
\label{sec:object_spatial_graph}
The object-based spatial graph $\mathcal{G}_s$ is constructed to capture the pairwise object interactions in each clip. We consider two spatial affinities to establish the graph: \textit{spatial relative arrangement} and 
\textit{object co-occurrence}. Hence, the spatial affinity matrix $\mathcal{E}_s$ consists of a position affinity matrix $\mathcal{E}^{pos}$ and a semantic affinity matrix $\mathcal{E}^{sem}$.
\\
\textbf{Position Affinity.} 
We use the mean IoU (mIoU)\footnote{For mIoU, we take the average of the IoU of two tubelet boxes in the overlapped duration.} of the box sequences of each clip tubelet pair to compute the position affinity. \textit{The diagonal values of $\mathcal{E}^{pos}$ is set to 1}.
\\
\textbf{Semantic Affinity.} We use the co-occurrence probability~\cite{chen2019knowledge} of the object categories of each clip tubelet pair as the semantic affinity, which is computed in the training set. Since the self co-occurrence is meaningless, \textit{the diagonal values of $\mathcal{E}^{sem}$ is set to 0}. 


\subsubsection{Relation-based Temporal Graph Construction}
\label{sec:relation_temporal_graph}
The relation-based temporal graph $\mathcal{G}_t$ is constructed to capture the temporal dependencies among adjacent relation instances. Since each node represents a tubelet pair, all possible connections in a video are tremendous.
As it is computationally expensive to operate on a large number of edges, we only keep the edges with high appearance affinity and location affinity. The assumption is, if two adjacent pairs look similar or have a smooth location transition, their relations are highly correlated.\\
\textbf{Appearance affinity.} For two tubelets $t_i$ and $t_j$ in adjacent clips, their appearance affinity value $\varphi ^{ap}$ is defined as the cosine similarity of their appearance feature $f^a$: 
\begin{equation}
     \varphi ^{ap}(t_i, t_j) = \begin{cases}
    \text{sim}(f^a(t_i), f^a(t_j)), & \varphi ^{ap} > \alpha \\ 
    0, & \text{otherwise}
    \end{cases},
\end{equation}
where $\alpha$ is the threshold to connect the clip tubelet pairs that have similar appearance features. \\
\textbf{Location affinity.} The location affinity value $\varphi ^{loc}$ of $t_i, t_j$  is defined as their volume IoU (vIoU), \ie, the Intersection Over Union of the spatial-temporal volume:
\begin{equation}
\begin{split}
\end{split}
    \varphi ^{loc}(t_i, t_j) = \begin{cases}
    \text{vIoU}(t_i, t_j), & \varphi ^{loc} > \beta \\ 
    0, & \text{otherwise}
    \end{cases}.
\end{equation}
where $\beta$ is the threshold to connect the clip tubelet pairs that are likely to belong to the same video trajectories.

\noindent\textbf{Temporal Affinity.} For two tubelet pairs in adjacent clips $(t_{si}, t_{oi})$ and $(t_{sj}, t_{oj})$ where $si$/$oi$ denotes the subject/object tubelet index, the temporal affinity $\mathcal{E}_t$ is the weighted\textbf{} sum of the appearance affinity and the location affinity: 
\begin{equation}
\begin{split}
    \mathcal{E}_t = & \lambda \min(\varphi ^{ap}(t_{si},t_{sj}),\varphi ^{ap}(t_{oi},t_{oj})) \\
    & + (1 - \lambda) \min(\varphi ^{loc}(t_{si},t_{sj}),\varphi ^{loc}(t_{oi},t_{oj})),
\end{split}
\end{equation}
where $\lambda$ is fusion weight. We choose the minimum affinity value of the subject and the object as the pair affinity value.

\subsubsection{Hierarchical Context Encoding}
Based on the object-based spatial graph and the relation-based temporal graph, we are able to build two types of contextual hierarchy to capture the spatial-temporal structure of the video, namely \textit{object-to-relation hierarchy} and \textit{spatial-to-temporal hierarchy}.
As shown in Figure~\ref{fig:arch}, we sequentially process the two types of graphs with two spatial GCNs and one temporal GCN.
The spatial GCNs extract low-level object-centric features from a single clip, and the temporal GCN captures high-level relation-centric features from continuous clips.
Both spatial GCNs and temporal GCN are implemented with the SAGEConv~\cite{hamilton2017inductive}.\\
\textbf{Parallel Spatial Graph Fusion.} To encode the graph heterogeneity in terms of object positional correlation and co-occur correlation, we use two GCNs to handle them in parallel. The clip-based appearance feature $F^a$ are the embeddings (\ie, messages) of the spatial graph nodes $V_s$, propagated with $\mathcal{E}^{pos}$ in position GCN and $\mathcal{E}^{sem}$ in semantic GCN (as adjacent matrix) to obtain corresponding positional features $F^{pos}$ and semantic features $F^{sem}$ respectively. 
Then a gate mechanism dynamically fuses the outputs as:
\begin{equation}
\begin{split}
  F^{pos} &= \text{SAGEConv}(F^a, \mathcal{G}_s^{pos}), \\
  F^{sem} &= \text{SAGEConv}(F^a, \mathcal{G}_s^{sem}), \\
  \mathbf{g} &= \text{sigmoid}(\mathbf{W_w}[F^{pos};F^{sem}]), \\
  F^{spa} &= \mathbf{g} \odot (\mathbf{W_f}[F^{pos};F^{sem}]),
\end{split}
\end{equation}
where $F^{spa}$ is the output of spatial GCNs, $\mathbf{W_w}$ and $\mathbf{W_f}$ are two learned linear transformations. $\odot$ indicates the element-wise multiplication operation, and $[;]$ denotes vector concatenation operation. \\
\textbf{Unidirectional Temporal Graph Reasoning.} Instead of using bidirectional node connections as in the spatial graphs, the messages in the temporal graph pass from the past relations pairs to the future relation pairs following the timeline.
The unidirectional message passing on the temporal graph aims to model the order of the events and alleviate the over-smoothing problem.
The clip-based relation features $f^{rel}$ of each clip tubelet pair are computed by concatenating the two object features $(f_i^{spa}, f_j^{spa})$ from the spatial graph outputs $F^{spa}$ and their relative spatial feature $f^{rp}$:
\begin{equation}
f^{rel} = \mathbf{W}_r[\mathbf{W}_s[(f_i^{spa};f_j^{spa})];f^{rp}]    
\end{equation}
where $\mathbf{W}_r, \mathbf{W}_s$ are learned linear transformations. \\
Then we feed the relation features of all clips $F^{rel}$ as temporal graph nodes $\mathcal{V}_t$ into the temporal GCN to obtain the higher-level relation features with edge weight $\mathcal{E}_t$, \ie,
\begin{equation}
\begin{split}
    F^{tem} &= \text{SAGEConv}(F^{rel}, \mathcal{G}_t)
\end{split}
\end{equation}
where $F^{tem}$ is the output of the temporal GCN. \\
\textbf{Discussion.}
To show the superiority of the proposed hierarchy from the object-based spatial graph to the relation-based temporal graph, we also explore some other possible spatial-temporal  graph architectures. More detailed experimental analysis and results are shown in Sec.~\ref{sec:ablation}.


\subsection{Relation Prediction}
\label{sec:relation_prediction}
The clip-level relations $R$ are predicted by classifying the multi-level spatial-temporal contextual features $F^{tem}$:
\begin{equation}
    R = \text{sigmoid}(\phi(F^{tem})),
\end{equation}
where $\phi(\cdot)$ is one MLP with two non-linear layers. During training, we use a Binary Focal loss (BFL) due to the class imbalance issue:
\begin{equation}
    \mathcal{L} = \mathcal{L}_{BFL}(R, \hat{R}),
\end{equation}
where $\hat{R}$ is the ground-truth predicate categories.

Then, we link all clip-level relations into video-level relations with greedy association algorithm~\cite{shang2017video}. 
In this paper, we relax the association condition that, if a relation can't find a association in the adjacent clip, one clip skip is allowed. The influence of the relaxed version is discussed in Sec.~\ref{sec:ablation}.

\section{Experiments.}
\label{sec:experiments}

\subsection{Experimental Settings} 
\noindent\textbf{Datasets.} We conducted experiments on two VidVRD benchmarks:
(1) \textbf{ImageNet-VidVRD}~\cite{shang2017video} contains $1,000$ videos with $132$ predicates and $35$ objects. The official splits are $800/200$ videos for training and test set.
(2) \textbf{VidOR}~\cite{shang2019annotating} contains $10,000$ videos with $50$ predicates and $80$ objects.
The official splits are $7,000/835/2,165$ videos for training, validation and test set. Since the test set is not released, we only evaluated HCM on the validation set.\\

\noindent\textbf{Evaluation Metric.}
We followed the prior works~\cite{liu2020beyond,chen2021social} to evaluate the performance of HCM on two common VidVRD tasks: 
(1) Relation Detection (\textbf{RelDet}): Predict the \texttt{subject-predict-object} triplet with spatial-temporal localization. 
The metrics are: mean Average Precision (\textbf{mAP}), Recall@50 (\textbf{R@50}) and Recall@100 (\textbf{R@100}).
(2) Relation Tagging (\textbf{RelTag}): Predict the \texttt{subject-predict-object} triplet. The metrics are Precision@1 (\textbf{P@1}), Precision@5 (\textbf{P@5}), Precision@10 (\textbf{P@5}). \\

\begin{table*}[tb]
\renewcommand\arraystretch{1.05}
    \centering
    \caption{Ablation experiments to assess the effectiveness of the spatial GCNs (S-GCNs, including one Pos-GCN and one Sem-GCN) and the temporal GCN (T-GCN) on the ImageNet-VidVRD dataset. We evaluate the individual contributions of each type of GCN by selectively adding them to the baseline model. }
    \label{tab:ablation_gcn}
        \begin{tabular}{C{1.4cm}C{1.4cm}C{1.4cm}|C{1.6cm}C{1.6cm}C{1.6cm}|C{1.6cm}C{1.6cm}C{1.6cm}}            \hline
            \multirow{2}{*}{Pos-GCN} & \multirow{2}{*}{Sem-GCN} & \multirow{2}{*}{T-GCN} & \multicolumn{3}{c|}{RelDet} & \multicolumn{3}{c}{RelTag}  \\
            & & &mAP & R@50 & R@100 & P@1 & P@5 & p@10  \\
            \hline
            & & & 25.68 & 14.40 & 16.38 & 72.00 & 53.20 & 39.80 \\
            \checkmark &  & &27.21 & 15.70 & 17.62 & 72.00 & 56.80 & 41.55 \\
             & \checkmark  & &27.12 & 15.16 & 17.64 & 70.00 & 56.60 & 40.70 \\
            \checkmark & \checkmark  & & 28.28 & 16.07 & 18.49 & 74.50 & 56.00 & 41.45 \\
            \checkmark & \checkmark  & \checkmark & \textbf{28.85} & \textbf{17.27} & \textbf{19.96} & \textbf{78.50} & \textbf{57.40} & \textbf{43.55} \\
            \hline
        \end{tabular}
\end{table*}

\noindent\textbf{Implementation Details.}
For the ImageNet-VidVRD dataset, we processed two released video-based features from BeyondShort~\cite{liu2020beyond} and BIG-C~\cite{gao2021classification} to obtain clip-based features (30 frames per clip with 15 frames overlapped) for a fair comparison. During training, we sampled consecutive clips from videos with a total size of 32 as a batch. We trained the model for 20 epochs with a learning rate drop by a factor of 10 after 10 epochs on one V100 GPU. $\lambda$ is set to 0.8. For the VidOR dataset, we followed BIG-C~\cite{gao2021classification} to generate the video-based features. The batch clip number is 64. 
We trained the model for 10 epochs with a learning rate drop by a factor of 10 after 5 epochs on one V100 GPU. $\lambda$ is set to 0.5.
On both datasets, the GCN layer number of the spatial GCNs and the temporal GCN is set to 3 and the hidden channels are 768. The activation function is ReLU. $\alpha = 0.8$, $\beta=0.7$. AdamW~\cite{loshchilov2017decoupled} optimizer is adopted.

\subsection{Ablation Studies} \label{sec:ablation}
We conducted ablation studies to demonstrate the effectiveness of each component in HCM and analyzed the key success factors of HCM as a clip-based framework.

\begin{table}[]
\renewcommand\arraystretch{1.05}
    \centering
    \caption{Results with different relation association settings on ImageNet-VidVRD}
    \label{tab:tab-vlink}
        \begin{tabular}{L{1.6cm}|C{1.6cm}C{1.6cm}C{1.6cm}}
         \hline
         \multirow{2}{*}{Models} & \multicolumn{3}{c}{RelDet}   \\
             &mAP & R@50 & R@100   \\
        \hline
        \textbf{HCM-vlink} & 25.73 & 11.27 & 12.29 \\
        \textbf{HCM-G} & 28.85 & 17.27 & 19.96 \\
        \textbf{HCM-RG} & \textbf{29.68} & \textbf{17.97} & \textbf{21.45} \\
        \hline
        \end{tabular}
\end{table}

\begin{table*}[tb]
\renewcommand\arraystretch{1.05}
\centering
\caption{Performance of the SOTA methods on ImageNet-VidVRD. Results in the same color (\textcolor{olive}{olive}, \textcolor{blue}{blue}, and \textcolor{red}{red}, except for black) adopt the same feature setting. 
M: relative motion (M* means BIG-C~\cite{gao2021classification} implicitly model it with query), C: object classeme. $\text{}^{\dagger}$ means the detector is finetuned on this dataset. $\text{}^{\ddagger}$ means the detector is trained on OpenImage~\cite{kuznetsova2020open}. VidVRD-II* is the reproduced results by us.}
\label{tab:tab-sota-vidvrd}
    \scalebox{1.0}{
    \begin{tabular}{L{2.6cm}|C{2.6cm}|C{2.6cm}|C{1.cm}C{1.cm}C{1.cm}|C{1.cm}C{1.cm}C{1.cm}}
        \hline
        \multirow{2}{*}{Models} & \multirow{2}{*}{Detector} & \multirow{2}{*}{Feature} & \multicolumn{3}{c|}{RelDet} & \multicolumn{3}{c}{RelTag}  \\
        & & &mAP & R@50 & R@100 & P@1 & P@5 & p@10  \\
        \hline
        \multicolumn{9}{l}{\emph{Clip-based Methods}} \\
        \hline
        VidVRD~\cite{shang2017video} & Faster-RCNN & iDT+M & 8.58 & 5.54 & 6.37 & 43.00 & 28.90 & 20.80 \\
        GSTEG~\cite{tsai2019video} & Faster-RCNN & iDT+M & 9.52 & 7.05 & 8.67 & 51.50 & 39.50 & 28.23 \\
        VRD-GCN~\cite{qian2019video} & Faster-RCNN & iDT+C+M & 16.26 & 8.07 & 9.33 & 57.50 & 41.00 & 28.50 \\
        MHA~\cite{su2020video} & Faster-RCNN & iDT+C+M & 19.03 & 9.53 & 10.38 & 57.50 & 41.40 & 29.45 \\ 
        IVRD~\cite{li2021interventional}& $\text{Faster-RCNN}^{\dagger}$ & RoI+M & 22.97 & 12.40 & 14.46 & 68.83 & 49.87 & 35.57\\
        VidVRD-II*~\cite{shang2021video}&$\text{Faster-RCNN}^{\ddagger}$ & RoI+M & \textcolor{olive}{23.09} & \textcolor{olive}{12.60} & \textcolor{olive}{14.13} & \textcolor{olive}{69.50} & \textcolor{olive}{52.50} & \textcolor{olive}{38.90}\\
        \hline
        \multicolumn{9}{l}{\emph{Video-based Methods}} \\
        \hline
        TSPN~\cite{woo2021and}&Faster-RCNN& RoI+C & 11.56 & 14.13 & 18.90 & 60.50 & 43.80 & 33.73 \\
        BeyondShort~\cite{liu2020beyond} & Faster-RCNN& RoI+I3D & 14.81 & 9.14 & 11.39 & 55.50 & 38.90 & 28.90 \\
        BeyondShort~\cite{liu2020beyond} & Faster-RCNN & RoI+I3D+M & \textcolor{blue}{18.38} & \textcolor{blue}{11.21} & \textcolor{blue}{13.69} & \textcolor{blue}{60.00} & \textcolor{blue}{43.10} & \textcolor{blue}{32.24} \\
        SocialFab~\cite{chen2021social} & Faster-RCNN & RoI+I3D+M & \textcolor{blue}{\textbf{ 19.77}} & \textcolor{blue}{12.91} & \textcolor{blue}{\textbf{16.32}} & \textcolor{blue}{61.00} & \textcolor{blue}{\textbf{47.50}} & \textcolor{blue}{\textbf{36.60}} \\
            BIG-C~\cite{gao2021classification} & Faster-RCNN & RoI+I3D+M* & 
            \textcolor{blue}{17.68} & \textcolor{blue}{9.70} & \textcolor{blue}{11.35} & \textcolor{blue}{56.00} & \textcolor{blue}{43.80} & \textcolor{blue}{32.85} \\
        BIG-C~\cite{gao2021classification} & MEGA & RoI+M* & \textcolor{red}{26.26} & \textcolor{red}{14.32} & \textcolor{red}{16.53} & \textcolor{red}{73.00} & \textcolor{red}{55.10} & \textcolor{red}{40.00} \\
        \hline
        \textbf{HCM-G} & $\text{Faster-RCNN}^{\ddagger}$ & RoI+M & \textcolor{olive}{\textbf{ 23.78}} & \textcolor{olive}{\textbf{12.89}} & \textcolor{olive}{\textbf{14.42}} & \textcolor{olive}{\textbf{75.00}} & \textcolor{olive}{\textbf{55.10}} & \textcolor{olive}{\textbf{41.65}} \\
        \textbf{HCM-G} & Faster-RCNN & RoI+I3D+M & \textcolor{blue}{18.41} & \textcolor{blue}{\textbf{13.01}} & \textcolor{blue}{15.47} & \textcolor{blue}{\textbf{65.00}} & \textcolor{blue}{44.00} & \textcolor{blue}{33.00}  \\
        \textbf{HCM-RG} & MEGA & RoI+M & \textcolor{red}{\textbf{29.68}} & \textcolor{red}{\textbf{17.97}} & \textcolor{red}{\textbf{21.45}} & \textcolor{red}{\textbf{78.5}} & \textcolor{red}{\textbf{57.4}} & \textcolor{red}{\textbf{43.55}} \\
        \hline
    \end{tabular}
    }
\end{table*}

\subsubsection{Effectiveness of HCM on Long-Term Relations} 
To verify the effectiveness of HCM on long-term relations with clip tubelets compared to directly using video tubelets, we tested HCM with the linkage from the video tubelets when associating the clip relations with video relations. In detail, since the clip tubelets used by HCM are generated by evenly partitioning the video tubelets used in \cite{gao2021classification} for a fair comparison, we linked the adjacent clips' relation triplets conditioning on that the subject and object belong to the same original video tubelets, denoting as \textbf{HCM-vlink}. 

As shown in Table~\ref{tab:tab-vlink}, HCM with greedy relation association~\cite{shang2017video} (\textbf{HCM-G}) reported better performance with a large margin than \textbf{HCM-vlink}. It proves that the partitioned clip tubelets are associated with more accurate subject/object trajectories than the adopted off-the-shelf video tracking algorithm. We thus verify that the relation clues can help reconnect the fragmented video tubelets to produce longer object trajectories. To further favor the long-term relation prediction, we also proposed a relaxed version of the greedy relation association (\textbf{HCM-RG}) as described in Sec.~\ref{sec:relation_prediction}, which further boosts the performance. Moreover, we find that the effect of RG is marginal (0.1\%) on VidOR compared to on ImageNetVidVRD (1.0\%) in terms of the average gains on RelDet. We infer that this is because VidOR is harder and has more long-term relation annotations.\\

\begin{table*}[tb]
\renewcommand\arraystretch{1.05}
\centering
\begin{minipage}{0.46\textwidth}
    \centering
    \caption{Ablations on the spatial affinity $\mathcal{E}_s$ and the temporal affinity $\mathcal{E}_t$ matrices by replacing it with adjacent matrix on ImageNet-VidVRD. }
    \label{tab:ablation-affinity}
            \begin{tabular}{L{1.4cm}|C{0.5cm}C{0.5cm}C{0.8cm}|C{0.5cm}C{0.5cm}C{0.8cm}}
             \hline
             \multirow{2}{*}{Models} & \multicolumn{3}{c|}{RelDet} & \multicolumn{3}{c}{RelTag}  \\
                 &mAP & R@50 & R@100 & P@1 & P@5 & P@10  \\
            \hline
            \textbf{HCM-G} & \textbf{28.85} & \textbf{17.27} & \textbf{19.96} & \textbf{78.50} & \textbf{57.40} & \textbf{43.55} \\
            \textbf{w/o $\mathcal{E}_s, \mathcal{E}_t$} & 28.31 & 16.07 & 18.22 & 77.00 & 55.90 & 39.15 \\
            \textbf{w/o $\mathcal{E}_t$} & 26.18 & 14.87 & 17.04 & 72.50 & 53.20 & 39.40 \\
            \textbf{w/o $\mathcal{E}_s$} & 28.26 & 16.88 & 19.36 & 77.00 & 56.50 & 42.50 \\
            \hline
            \end{tabular}
\end{minipage}
\hspace{0.04\linewidth}
\begin{minipage}{0.46\textwidth}
    \centering
    \caption{Results of the other possible spatial-temporal architectures on ImageNet-VidVRD.}
    \label{tab:ablation-hierar}
        \scalebox{1.0}{
            \begin{tabular}{L{1.4cm}|C{0.5cm}C{0.5cm}C{0.8cm}|C{0.5cm}C{0.5cm}C{0.8cm}}
             \hline
             \multirow{2}{*}{Models} & \multicolumn{3}{c|}{RelDet} & \multicolumn{3}{c}{RelTag}  \\
                &mAP & R@50 & R@100 & P@1 & P@5 & P@10 \\
            \hline
            \textbf{HCM-G} & \textbf{28.85} & \textbf{17.27} & \textbf{19.96} & \textbf{78.50} & \textbf{57.40} & \textbf{43.55} \\
            \textbf{parallel} & 28.54 & 16.67 & 19.05 & 75.50 & 57.00 & 43.00 \\
            \hline
            \textbf{w/o $\mathcal{E}$} & \textbf{28.31} & \textbf{16.07} & \textbf{18.22} & \textbf{77.00} & \textbf{55.90} & 39.75 \\
            \textbf{reverse} & 27.25 & 15.57 & 18.22 & 74.50 & 54.20 & \textbf{40.00} \\
            \textbf{pure-obj} & 26.18 & 14.87 & 17.04 & 72.50 & 53.20 & 39.40 \\
            \hline
            \end{tabular}
        }
\end{minipage}
\end{table*}

\subsubsection{Component Analysis of Hierarchical Context} In order to verify the effectiveness of spatial affinity graphs and temporal affinity graphs in the hierarchical context encoding. Firstly, we analyzed the impact of the three GCNs on performance. 
Secondly, we analyzed the effectiveness of using the spatial affinity and the temporal affinity in GCNs as edge weights. \\

\noindent(a) \textit{Impact of Spatial GCNs (Pos-GCN and Sem-GCN) and Temporal GCN.}
The baseline (\textbf{Base}) predicts the relations without hierarchical context encoding.
Then we added each GCN to the baseline to evaluate the performance gains.
As shown in Table~\ref{tab:ablation_gcn}, the Pos-GCN (\textbf{Base+Pos-GCN}) and the Sem-GCN (\textbf{Base+Sem-GCN}) brings $1.36\%/1.78\%$ and $1.15\%/0.77\%$ average gains on RelDet/RelTag respectively. 
We observed that the fusion of Pos-GCN and Sem-GCN (\textbf{Base+S-GCNs}) shows better results with $2.13\%/2.32\%$ average gains on RelDet/RelTag.
Finally, applying the temporal GCN (\textbf{Base+S-GCNs+T-GCN}) further brings $1.08\%/2.50\%$ average gains on RelDet/RelTag.\\

\noindent(b) \textit{Effectiveness of Spatial and Temporal Affinities.} The baseline uses the default edge function (unweighted) of SAGE in the three GCNs (\textbf{HCM-G w/o $\mathcal{E}_s, \mathcal{E}_t$}) and the temporal graph will be a dense graph without $\mathcal{E}_t$.
As shown in Table~\ref{tab:ablation-affinity}, 
since using the dense temporal graph is bad for training, we can see that adding the spatial affinities (\textbf{HCM-G w/o $\mathcal{E}_t$}) can do more harm than good compared with the baseline (\textbf{HCM-G w/o $\mathcal{E}_s$,$\mathcal{E}_t$}).
And when using the temporal affinities (\textbf{HCM-G w/o {$\mathcal{E}_s$}}) to ensure a sparse temporal graph, the performance is approximate to the baseline (\textbf{HCM-G w/o $\mathcal{E}_s$,$\mathcal{E}_t$}).
Using both affinities (\textbf{HCM-G}) can lead to performance gains on all metrics (e.g., 1.5\% on RelTag P@1).
Hence, spatial affinity and temporal affinity work when applied jointly. \\

\noindent(c) \textit{Unidirectional Temporal Reasoning.} To verify the effectiveness of the unidirectional temporal graph according to the timeline, we also experimented with bidirectional reasoning that each relation node sends messages to both the previous and the next clip (\textbf{BI-HCM}) and inverted unidirectional reasoning (\textbf{IN-HCM}) that each relation node only sends messages to the previous clip. As shown in Table~\ref{tab:tab-direct}, both bidirectional message passing and inverted message passing causes a performance drop of all metrics on VidOR.\\

\begin{table}
\renewcommand\arraystretch{1.05}
    \centering
    \caption{Ablation study on the effectiveness of the Binary Focal Loss (BFL) on ImageNet-VidVRD. }
    \label{tab:ablation-bfl}
            \begin{tabular}{L{1.4cm}|C{0.5cm}C{0.5cm}C{0.8cm}|C{0.5cm}C{0.5cm}C{0.8cm}}
             \hline
             \multirow{2}{*}{Models} & \multicolumn{3}{c|}{RelDet} & \multicolumn{3}{c}{RelTag}  \\
                 &mAP & R@50 & R@100 & P@1 & P@5 & P@10  \\
            \hline
            HCM-BCE & 27.06 & 15.04 & 17.08 & 77.00 & 55.90 & 40.80 \\
            HCM-BFL & \textbf{28.85} & \textbf{17.27} & \textbf{19.96} & \textbf{78.50} & \textbf{57.40} & \textbf{43.55} \\
            \hline
            \end{tabular}
\end{table}

\subsubsection{Ablations on Hyper-Parameter Designs} We conducted extensive ablation studies on some hyper-parameter settings.\\
\noindent(a) \textit{Effectiveness of Binary Focal Loss (\textbf{BFL}).} 
Please note that, in our experiments, we utilized BFL in the reproduced results of the state-of-the-art clip-based method VidVRD-II*, as shown in Table~\ref{tab:tab-sota-vidvrd}. 
To ensure fair comparisons with the video-based methods and show the effectiveness of BFL, we also reported the results of HCM-G with BCE Loss on the ImageNet-VidVRD dataset, as shown in Table~\ref{tab:ablation-bfl}. 
While the performance without using BFL experiences a drop, HCM with BCE Loss still achieves the best overall performance.
It is worth noting that clip-based methods face a more significant imbalance issue compared with video-based methods. This arises from the nature of encoding relations between pairwise objects in multiple clips. 
\\

\noindent(b) \textit{Affinity Thresholds for Temporal Graph Connections.}
We examined the impact of different threshold settings for the temporal affinity ($\alpha$ and $\beta$) in HCM-G. The results are presented in Table~\ref{tab:ablation-alpha-beta}. It is evident that setting $\beta > 0.7$ leads to performance plateaus, indicating the significance of the temporal affinity in reducing the connectivity complexity in the temporal graph. Moreove, adjusting the $\alpha$ threshold primarily affects the RelDet metrics, resulting in a performance drop.
\\

\noindent(c) \textit{Fusion Weight of Appearance and Location Affinities.}
We conducted experiments to evaluate the performance using different fusion weights ($\lambda$) for appearance affinity and location affinity. The results are summarized in Table~\ref{tab:ablation-lambda}. It can be observed that a fusion weight of $\lambda = 0.8$ achieves the optimal performance, while the performance remains relatively similar for other fusion settings. \\

\begin{table*}[tb]
\renewcommand\arraystretch{1.05}
\centering
\caption{Performance of the SOTA methods on VidOR. L: word2vec language feature.}
\label{tab:tab-sota-vidor}
    \begin{tabular}{L{2.5cm}|C{2.5cm}|C{2.5cm}|C{0.9cm}C{0.9cm}C{0.9cm}|C{0.9cm}C{0.9cm}C{0.9cm}}
        \hline
        \multirow{2}{*}{Models} & \multirow{2}{*}{Detector} & \multirow{2}{*}{Feature} & \multicolumn{3}{c|}{RelDet} & \multicolumn{3}{c}{RelTag}  \\
        & & &mAP & R@50 & R@100 & P@1 & P@5 & p@10  \\
        \hline
        \multicolumn{9}{l}{\emph{Clip-based Methods}} \\
        \hline
         MHA~\cite{su2020video} & FGPA & M+L & 6.59 & 6.35 & 8.05 & 50.72 & 41.56 & - \\
        IVRD~\cite{li2021interventional}&Faster-RCNN&RoI+M & 7.42 & 7.36 & 9.41 & 53.40 & 42.70 & -\\
        \hline
        \multicolumn{9}{l}{\emph{Video-based Methods}} \\
        \hline
        TSPN~\cite{woo2021and}&Faster-RCNN& RoI+C & 7.61 & 9.33 & 10.71 & 53.14 & 42.22 & 34.94 \\
        BeyondShort~\cite{liu2020beyond} & RefineDet & RoI+I3D+M & 6.85 & 8.21 & 9.90 & 51.20 &  40.73 & - \\
        SocialFab~\cite{chen2021social} & Faster-RCNN & RoI+I3D+L+M & 10.04 & 8.94 & 10.69 & 61.52 & 50.05 & 38.48 \\
        BIG-C~\cite{gao2021classification} & MEGA & RoI+M* & \textcolor{red}{8.03} & \textcolor{red}{7.61} & \textcolor{red}{9.40} & \textcolor{red}{62.25} & \textcolor{red}{50.91} & \textcolor{red}{40.27} \\
        BIG-C~\cite{gao2021classification} & MEGA & RoI+I3D+M*+L & 9.35 & 8.79 & 11.30 & 64.30 & 52.35 & 41.85 \\
        \hline
        \textbf{HCM-RG} & MEGA & RoI+M & \textcolor{red}{\textbf{10.44}} & \textcolor{red}{\textbf{9.74}} & \textcolor{red}{\textbf{11.23}} & \textcolor{red}{\textbf{67.43}} & \textcolor{red}{\textbf{52.19}} & \textcolor{red}{\textbf{40.30}} \\
        \hline
    \end{tabular}
\end{table*}

\subsubsection{Ablations on the Spatial-Temporal Architecture} 
We do ablation studies on three features of HCM: (i) the hierarchical architecture, (ii) the spatial-to-temporal graph connections and (iii) the object-to-relation graph nodes.

\begin{figure*}[!tb]
        \centering
        \includegraphics[width=0.9\linewidth]{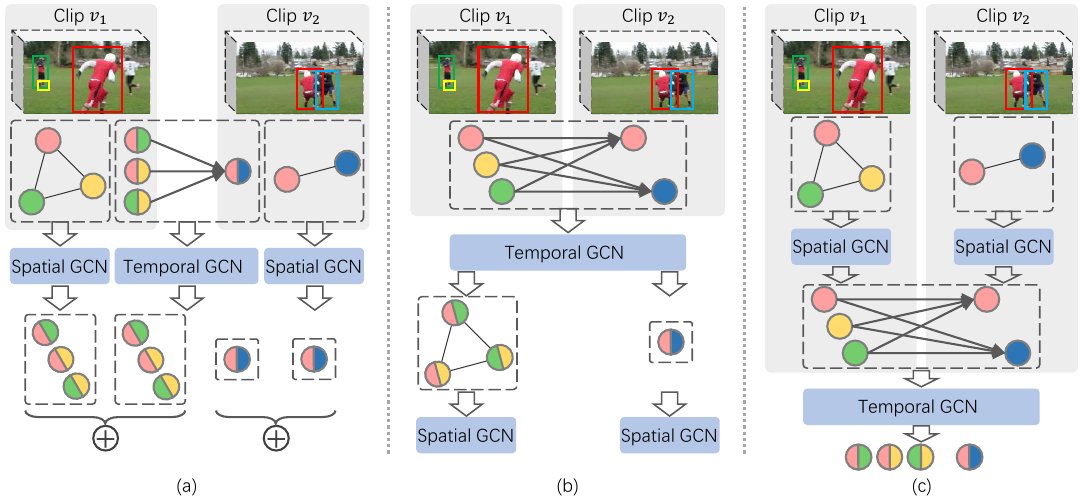}
        \caption{Ablation experiments on the three different versions of the spatial-temporal graph structures: (a) \textbf{Parallel Architecture}: the message passing occurs first on the relation-based temporal graph and then on the object-based spatial graph. (b) \textbf{Reversed Architecture}: the message passing occurs firstly on the relation-based temporal graph and then the object-based spatial graph. (c) \textbf{Pure-Object based Architecture}: the graph nodes of the temporal graph are the same object nodes as the spatial graph. This means that the temporal graph represents the temporal relationships between objects rather than object relation pairs. We use the two adjacent clips to illustrate the spatial-temporal context encoding process.}
        \label{fig:arch_abl}
        
\end{figure*}

\noindent(a) \textbf{\textit{Hierarchical vs. Parallel.}}
To show the effectiveness of the hierarchical architecture of the spatial GCNs and the temporal GCN, we designed a parallel architecture (\textbf{parallel}) as shown in Figure~\ref{fig:arch_abl} (a) which propagates messages on the spatial and temporal graphs in parallel and then fuse the outputs for relation prediction.
As shown in Table~\ref{tab:ablation-hierar}, this parallel version (\textbf{parallel}) also achieved the state-of-the-art performance but is inferior on RelDet (all) and RelTag (P@1) thus verifying the superiority of the hierarchical architecture.\\

\noindent(b) \textbf{\textit{Spatial-to-Temporal vs. Temporal-to-Spatial.}}
To show the effectiveness of connecting the objects in the spatial dimension and the relations in the temporal dimension, we designed a reversed version of HCM's architecture as shown in Figure~\ref{fig:arch_abl} (b) that connecting the objects in the temporal dimension and the relations in the temporal dimension. To eliminate the influence of the spatial and temporal affinities in HCM, we used \textbf{HCM-G w/o $\mathcal{E}$} for a fair comparison. As shown in Table~\ref{tab:ablation-hierar}, the reversed version (\textbf{reverse}) shows inferior performance on most of the metrics.\\

\noindent(c) \textbf{\textit{Object-to-Relation vs. Object-to-Object.}}
To show the importance of the high-level temporal connections based on relationships, we designed a pure object based spatial-temporal architecture as shown in Figure~\ref{fig:arch_abl} (c) that changes the graph nodes in the temporal graph from tubelet pairs to tubelets (i.e., same as the spatial graph nodes). Similarly we use \textbf{HCM-G w/o $\mathcal{E}$} for a fair comparison. As shown in Table~\ref{tab:ablation-hierar}, this pure object version (\textbf{pure-obj}) causes a significant performance drop on all metrics. 

\begin{table}
\renewcommand\arraystretch{1.05}
    \centering
    \caption{Results of HCM-G with different choices of $\alpha$ and $\beta$ on ImageNet-VidVRD. }
    \label{tab:ablation-alpha-beta}
            \begin{tabular}{L{0.6cm}L{0.6cm}|C{0.5cm}C{0.5cm}C{0.8cm}|C{0.5cm}C{0.5cm}C{0.8cm}}
             \hline
             \multirow{2}{*}{$\alpha$} & \multirow{2}{*}{$\beta$} & \multicolumn{3}{c|}{RelDet} & \multicolumn{3}{c}{RelTag}  \\
                & &mAP & R@50 & R@100 & P@1 & P@5 & P@10  \\
            \hline
            0.8 & 0.6 & 27.79 & 16.65 & 19.30 & 76.00 & 57.00 & 43.15 \\
            0.8 & 0.7 & 28.85 & 17.27 & \textbf{19.96} & \textbf{78.50} & 57.40 & \textbf{43.55} \\
            0.8 & 0.8 & 28.04 & 16.90 & 19.46 & 74.50 & 57.70 & 43.00 \\
            0.8 & 0.9 & 28.08 & 17.15 & 19.65 & 74.50 & 57.80 & 43.10 \\
            0.7 & 0.7 & 28.00 & 17.02 & 19.50 & 78.00 & 57.20 & 43.55 \\
            0.9 & 0.7 & 27.86 & 16.63 & 19.11 & 74.00 & \textbf{57.90} & 43.20 \\
            \hline
            \end{tabular}
\end{table}

\begin{table}
\renewcommand\arraystretch{1.05}
    \centering
    \caption{Results of HCM-G with different choices of $\lambda$ on ImageNet-VidVRD. }
    \label{tab:ablation-lambda}
            \begin{tabular}{L{1.0cm}|C{0.5cm}C{0.5cm}C{0.8cm}|C{0.5cm}C{0.5cm}C{0.8cm}}
             \hline
             \multirow{2}{*}{$\lambda$} & \multicolumn{3}{c|}{RelDet} & \multicolumn{3}{c}{RelTag}  \\
              &mAP & R@50 & R@100 & P@1 & P@5 & P@10  \\
            \hline
            0.9 & 28.04 & 16.90 & 19.46 & 74.50 & \textbf{57.50} & 42.90 \\
            0.8 & \textbf{28.85} & \textbf{17.27} & \textbf{19.96} & \textbf{78.50} & 57.40 & \textbf{43.55} \\
            0.7 & 28.16 & 17.02 & 19.17 & 74.50 & 56.80 & 42.70 \\
            0.5 & 28.18 & 17.10 & 19.44 & 76.50 & 56.20 & 42.85 \\
            \hline
            \end{tabular}
\end{table}

\subsection{Qualitative Results.} 
\begin{figure*}
    \centering
    \includegraphics[width=1.0\linewidth]{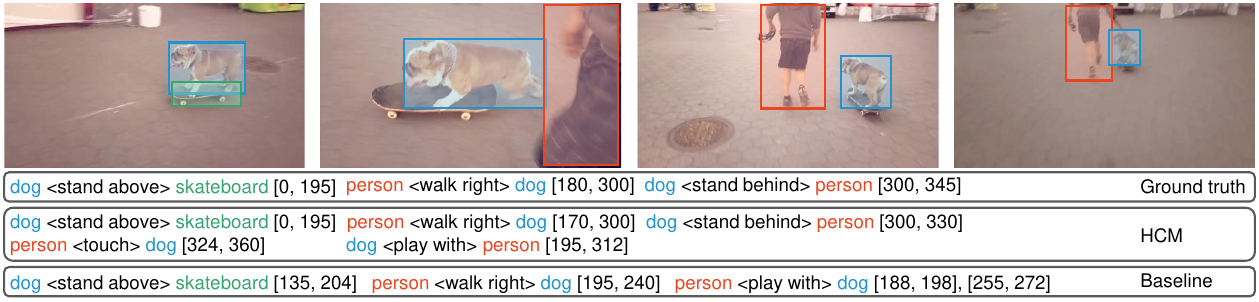}
    \caption{Qualitative results on the ImageNet-VidVRD dataset, showcasing relation triplets with their corresponding durations represented as frame indexes within square brackets. The results are presented in three solid boxes, arranged from top to bottom, representing the Ground Truth: the actual relation triplets annotated in the dataset; HCM (Hierarchical Context Model): the relation triplets predicted by our proposal method which accurately capture the temporal dynamics of the relations, providing relation triplets with precise durations that closely match the ground truth annotations.; the Baseline model: the relation triplets predicted by the baseline model which exhibit limitations in accurately capturing the temporal aspects of the relations.}
    \label{fig:quality}
\end{figure*}

In Figure~\ref{fig:quality}, we present the qualitative results from our experiments on the ImageNet-VidVRD dataset, comparing the performance of our Hierarchical Context Model (HCM) with a baseline model.
We observe that HCM accurately predicts the relations with precise temporal boundaries, capturing both short-term and long-term relationships. It successfully identifies true relations that were not annotated, such as the short-term relation ``\texttt{touch}'' which relies on fine-grained spatial cues, and the long-term relation ``\texttt{play with}'' which requires temporal reasoning. In contrast, the baseline model predominantly predicts relations with short-term temporal boundaries and lacks effective temporal modeling. It fails to recognize the relation ``\texttt{stand behind}'' altogether. Even though the baseline model can predict the long-term relation ``\texttt{play with}'' it erroneously breaks it into two separate segments.
The results highlight the significance of modeling the spatial-temporal structure within the clip-based pipeline. By incorporating spatial and temporal context at both object and relation level, HCM is able to produce more accurate short-term and long-term predictions. 

\begin{table}[]
\renewcommand\arraystretch{1.05}
    \centering
    \caption{Results on different message passing directions of temporal reasoning on VidOR: \textbf{BI-HCM} refers to the bidirectional message passing on the temporal graph; \textbf{IN-HCM} refers to the inverted unidirectional message passing (from future to past) on the temporal graph.}
        \label{tab:tab-direct}
            \begin{tabular}{L{1.5cm}|L{1.5cm}|L{2.cm}|L{2.cm}}
                \hline
                \multicolumn{2}{l|}{Models} & \textbf{BI-HCM} & \textbf{IN-HCM} \\
                \hline
                \multirow{3}{*}{{RelDet}} & mAP & 9.33 $_{\color{red}{-1.11}}$ & 9.44 $_{\color{red}{-1.00}}$\\
                 & R@50 & 9.21 $_{\color{red}{-0.53}}$ & 9.18 $_{\color{red}{-0.56}}$ \\
                 & R@100 & 10.63 $_{\color{red}{-0.60}}$ & 10.64 $_{\color{red}{-0.59}}$ \\
                \hline
                \multirow{3}{*}{{RelTag}} & P@1 & 65.02 $_{\color{red}{-2.41}}$ & 63.94 $_{\color{red}{-3.49}}$ \\
                 & P@5 & 49.45 $_{\color{red}{-2.74}}$ & 50.36 $_{\color{red}{-1.83}}$ \\
                 & P@10 & 38.04 $_{\color{red}{-2.26}}$ & 38.98 $_{\color{red}{-1.32}}$  \\
                \hline
            \end{tabular}
\end{table}
 

\subsection{Comparisons with State-of-the-Arts}
\subsubsection{Settings} We compared our method with the state-of-the-art methods on both Imagenet-VidVRD and VidOR as shown in Table~\ref{tab:tab-sota-vidvrd} and Table~\ref{tab:tab-sota-vidor} respectively. We divided them to the clip-based methods (\textbf{VidVRD}~\cite{shang2017video},\textbf{GSTEG}\cite{tsai2019video}, \textbf{VRD-GCN}~\cite{qian2019video}, \textbf{MHA}~\cite{su2020video}  \textbf{IVRD}~\cite{li2021interventional}, \textbf{VidVRD-II}~\cite{shang2021video}) and the video-based methods (\textbf{\textbf{TSPN}~\cite{woo2021and}, \textbf{BeyondShort}\cite{liu2020beyond}}, \textbf{SocialFab}~\cite{chen2021social},\textbf{BIG-C}~\cite{gao2021classification}). Since the existing methods used different object detectors finetuned with various datasets and different feature combinations (\eg, RoI, I3D, motion and language features), it's challenging to conduct comprehensive comparison with all features.
We reported the results using three types of released data: \textbf{VidVRD-II}~\cite{shang2021video}, \textbf{BeyondShort}~\cite{liu2020beyond} and \textbf{BIG-C}~\cite{gao2021classification} for fair comparison.\\

\subsubsection{Results on ImageNet-VidVRD} 
As shown in Table~\ref{tab:tab-sota-vidvrd}, HCM achieved a new state-of-the-art performance using the same data settings of \cite{shang2021video} (shown in \textcolor{olive}{olive}, \eg, $0.69\%$ on RelDet mAP, $5.5\%$ on RelTag P@1) and BIG-C~\cite{gao2021classification} (shown in \textcolor{red}{red}, \eg, $2.59\%$ on RelDet mAP, $5.5\%$ on RelTag P@1). 
When using the data of \cite{liu2020beyond} (shown in \textcolor{blue}{blue}), HCM achieved the best performance on RelDet R@50, and RelTag P@1, and the second best on the other metrics. 
We also found that HCM showed more significant performance gains when the tracking result is more accurate (\textcolor{red}{red} $>$ \textcolor{olive}{olive} $>$ \textcolor{blue}{blue}).\\

\subsubsection{Results on VidOR} 
When evaluating on the VidOR dataset, we observed that the choice of tubelet generation methods had a weaker impact compared to the ImageNet-VidVRD dataset. Table~\ref{tab:tab-sota-vidor} presents the results, showcasing the superiority of our HCM across all evaluation metrics.
Specifically, in comparison to the video-based method BIG-C~\cite{gao2021classification} under the same feature setting (shown in \textcolor{red}{red}), HCM consistently outperforms it in terms of all metrics. 
Furthermore, even when utilizing only the RoI and relative motion features, HCM still achieves highly competitive results when compared to methods that leverage more powerful tubelet features such as I3D features and word2vec language features. 

\section{Conclusion}
\label{sec:conc}
In this paper, we introduced the Hierarchical Context Model (HCM) as a solution for short-term and long-term visual relation detection in videos.
HCM design object-based spatial graphs and relation-based temporal graphs hierarchically to effectively model multi-level spatial correlations among objects and high-level temporal correlations among relation pairs. 
Our HCM offers flexibility and scalability in handling various durations and context levels in VidVRD.
Through extensive experiments on two widely-used benchmarks (i.e. ImageNet-VidVRD and VidOR), we validated the effectiveness of each component of HCM and the unique spatial-temporal graph design. 
However, we acknowledge some limitations of our approach. Firstly, the dense short clips in HCM can lead to redundant relation predictions, which may impact efficiency. Secondly, when there are drastic changes in video lengths, the training of HCM may become unstable.
In future work, we plan to explore more reasonable clip sampling methods and robust graph batching strategies to enhance stability and adaptability to varying video lengths.
Additionally, it is crucial to consider ethical implications, as HCM's ability to track people and understand their relations with surrounding objects raises privacy concerns. We will ensure that privacy protection measures are implemented and privacy implications are carefully addressed in future research.

{
\bibliographystyle{IEEEtran}
\bibliography{egbib}
}

\end{document}